\documentclass[sigconf,authorversion,nonacm]{acmart}
\usepackage{array,multirow,graphicx}
\usepackage{makecell}
\hypersetup{
    colorlinks=true,
    linkcolor=blue,
    filecolor=red,      
    urlcolor=magenta,
    breaklinks=true,            
}
\usepackage{breakurl}  

\begin{document}
\title{Summarisation of German Judgments in conjunction with a Class-based Evaluation}

\author{Bianca Steffes}
\affiliation{%
  \institution{Saarland University, Saarland Informatics Campus, Chair of Legal Informatics}
  \city{Saarbrücken}
  \country{Germany}
  }
\email{bianca.steffes@uni-saarland.de}

\author{Nils Torben Wiedemann}
\affiliation{%
  \institution{Saarland University, Saarland Informatics Campus, Chair of Legal Informatics}
  \city{Saarbrücken}
  \country{Germany}
  }
\email{nils_torben.wiedemann@uni-saarland.de}

\author{Alexander Gratz}
\affiliation{%
  \institution{Rechtsanwälte Zimmer-Gratz}
  \city{Bous}
  \country{Germany}
  }
\email{mail@zimmer-gratz.de}

\author{Pamela Hochreither}
\affiliation{%
  \institution{Saarland University, Chair of Civil Law and Civil Procedure Law}
  \city{Saarbrücken}
  \country{Germany}
  }
\email{pamela.hochreither@uni-saarland.de}

\author{Jana Elina Meyer}
\affiliation{%
  \institution{Saarland University, Chair of Civil Law and Civil Procedure Law}
  \city{Saarbrücken}
  \country{Germany}
  }
\email{jana_elina.meyer@uni-saarland.de}

\author{Katharina Luise Schilke}
\affiliation{%
  \institution{Saarland University, Saarland Informatics Campus, Chair of Legal Informatics}
  \city{Saarbrücken}
  \country{Germany}
  }
\email{katharina.schilke@uni-saarland.de}

\renewcommand{\shortauthors}{Steffes et al.}

\begin{abstract}
 The automated summarisation of long legal documents can be a great aid for legal experts in their daily work. We automatically create summaries (\textit{guiding principles}) of German judgments by fine-tuning a decoder-based large language model. We enrich the judgments with information about legal entities before the training. For the evaluation of the created summaries, we define a set of evaluation classes which allows us to measure their language, pertinence, completeness and correctness. Our results show that employing legal entities helps the generative model to find the relevant content, but the quality of the created summaries is not yet sufficient for a use in practice.\footnote{This paper is an extended preprint of the paper of the same title published at ICAIL 2025.}
\end{abstract}



\maketitle

\section{Introduction} 
The legal field is notorious for long and complex documents whose key points can only be grasped by investing a substantial amount of time to read (and comprehend) them. The automated summarisation of these long documents can considerably reduce this amount of time and thus is of particular importance for the legal domain. Existing research in this field is mainly conducted using English texts (e.g., \cite{Luhn58, Nallapati&17SummaRunner, Liu&19BertSum}) whilst other languages like German 
are rarely covered. 

In our work, we extend the comparatively small body of research on summarising German legal documents with a particular emphasis on judgments. Our main objective is the automated creation of so-called \textit{guiding principles} for (German) judgments. The overarching concept of guiding principles could serve as a role model for the summary of extensive and complex legal documents across legal domains and legal systems. The main advantage of focusing on German guiding principles is the rather common practice of the higher German courts to write such guiding principles, providing a sufficient amount of publicly available data for research purpose. 
We generate guiding principles by fine-tuning a state of the art decoder-based language model -- once on the original judgments and once after enriching the judgments used as training data with information about the legal entities proposed by Leitner et al. \cite{Leitner&20Entities}. We examine the quality of the created summaries through an application of both automated evaluation metrics (ROUGE and BERTScore) and specifically defined evaluation classes. Our results show that the detection of relevant content by the generative model benefits from the enriched judgments.

For the purpose of this contribution, we review the relevant related research (Section \ref{sec:related_work}) and specify the summarisation task to be solved, as well as the employed language model (Section \ref{sec:task}). In the ensuing section (Section \ref{sec:classes}) we present the evaluation classes for the assessment of generated summaries, discuss their relation to pre-existing evaluation methods and show a proof of concept application of the classes. In Section \ref{sec:round_2} we then apply these classes to the results created by a fine-tuned state of the art decoder-based language model and show possible improvements using legal entities. We conclude our paper with the discussion of our results (Section \ref{sec:discussion}) and an outlook for future work (Section \ref{sec:conclusion}).

\section{Related Work} \label{sec:related_work}
A summary can be considered a shortened version of a document containing only the information relevant for a specific audience for a certain objective. Summaries can be classified into \textit{extractive} and \textit{abstractive} summaries. Extractive summaries encompass summaries comprised of sentences extracted from the document. Abstractive summaries, on the other hand, are summaries that paraphrase the relevant content of the document and thus can contain words not included in the summarised text. 

The automated summary of (legal) documents has been the focus of various scientific contributions. Initial research, like the seminal work of Luhn \cite{Luhn58}, who creates extractive summaries based on word frequencies and positions, focuses on creating extractive summaries by selecting sentences according to a calculated relevance score. Erkan et al. \cite{Erkan&LexRank04} present \textit{LexRank} which calculates the relevance (in their case: \textit{centrality}) of a sentence using a graph-based approach. The sentences are represented as nodes with edges between sentences of a certain similarity. The overall relevance of a sentence is based on its edges and can be calculated with different centrality measures. Yeh et al. \cite{Yeh&05LSA} propose two different approaches for extractive summarisation. In their modified corpus-based approach they provide a trainable summariser which calculates the relevance of sentences using sentence features like the position, contained keywords or the resemblance to the title. In their second algorithm they use latent semantic analysis in combination with a text relationship map to rank and select sentences. He et al. \cite{He&Reconstruction12} try to solve the summarisation task starting from a different point of view. They argue that a summary represents the original document best if it allows the reconstruction of large parts of the original document. Moawad et al. \cite{Moawad&RedSum12} generate abstractive summaries by using a graph based method. They first create rich semantic graphs, reduce these graphs and then generate the text of the summary from the resulting graphs. 

Similar concepts can also be applied to the legal domain. Farzindar et al. \cite{Farzindar&LetSum04} create extractive table-style summaries of Canadian judgments. They divide the original judgment into thematic segments and select relevant sentences to these parts based on sentence features like word occurrences or position in the judgment. Hachey et al. \cite{Hachey&ExtractiveLegal06} generate extractive summaries of judgments of the UK House of Lords. They assign rhetorical roles to sentences and combine these roles with their calculated relevance to select the sentences for the summaries. Kim et al. \cite{Kim&CohesionGraph13} use a graph-based approach to create extractive summaries for the same data set. They convert the documents to be summarised into directed graphs with asymmetric weights and choose sentences of a high cohesion for their summaries. The cohesion between two sentences is represented as the likelihood that one sentence is embedded in the other. Polsley et al. \cite{Polsley&CaseSummarizer16} generate extractive summaries of Australian judgments. They rank sentences using structural features of the document and also provide a user interface for their work. Bhattacharya et al. \cite{Bhattacharya&DelSumm21} create extractive summaries of Indian judgments. They use Integer Linear Programming to optimise the parameters of a hand crafted ranking function based on specific aspects of the judgments that a legal professional would also pay attention to. 

In recent years, the advancements in the field of neural networks and artificial intelligence lead to considerable improvements in solving summarisation tasks and abstractive summarisation in particular. Nallapati et al. \cite{Nallapati&17SummaRunner} apply Gated Recurrent Neural Networks to write extractive summaries. Their approach for the creation of these summaries is based on a binary classification of every sentence -- either the sentence is contained in the extractive summary or not. Liu et al. \cite{Liu&19BertSum} apply a similar approach but use BERT for the classification. For extractive summaries, they separate the sentences by inserting CLS tokens between them and derive the classification output for each sentence from the output of the corresponding CLS token. For abstractive summarisation, they propose the use of a combination of BERT as an encoder and a (randomly initialized) decoder for the output generation. Liu et al. \cite{Liu&GIST19}  apply this classification task to the legal field and create extractive summaries of Chinese judgments. They use gradient boosting and different types of neural networks to classify whether a sentence should be contained in the summary or not. Zhong et al. \cite{Zhong&MMR19} assume that sentences which are predictive for a judgment's outcome are best suited for a summary. Using a Convolutional Neural Network, they repeatedly choose the most predictive sentence for the summary (always removing the already chosen sentences from all choosable sentences) until a fixed length of the summary is reached. Xu et al. \cite{Xu&21SumExtract} propose that the identification of \textit{legal argument triples} might be helpful for selecting suitable sentences for a summary. These triples each consist of the issue, the conclusion and the reasons for a specific legal point of discussion. Yoon et al. \cite{Yoon&22LLMKorea} generate abstractive summaries using pre-trained large language models. They fine-tune the Korean variants of BART and BERT2BERT to summarise Korean judgments. Nguyen et al. \cite{Nguyen&23KeyWords} create abstractive summaries of US legal cases using different Transformer encoder-decoder models. They preselect sentences by the occurrence of certain keywords and generate the summary from these selected sentences. 

Although a lot of research on summarisation exists on an international scale, research on the summarisation of German texts is comparatively scarce and under-represented
(Aumiller et al. \cite{Aumiller&23SotA} give a good overview). Elnaggar et al. \cite{Elnaggar&18MultiTask} use multi-task learning with the MultiModel algorithm to summarise documents of the European Parliament in different languages including German. Moro et al. \cite{Moro&24Multi-Lingual} show how to use multi-language transfer learning on Australian Legal Case Report. They translate the judgments into different languages -- including German -- and create abstractive summaries with a multilingual model. Schubiger \cite{Schubiger24GermanSummarizationLLMs} generate abstractive summaries of German texts by prompting different pre-trained large language models. He evaluates the effectiveness of different prompting methods for summarising German news or Wikipedia articles. 
There is also some work on native German legal texts. Steffes et al. \cite{Steffes&22ArgStruct} create extractive summaries of judgments of the German Federal Court of Justice. They rank sentences based on structural aspects of the document (e.g., word or sentence positions) and word frequencies. Glaser et al. \cite{Glaser&21German} apply different variations of neural networks to generate extractive and abstractive summaries for judgments of various courts. 

Regarding the evaluation of generated summaries, ROUGE (Recall-Oriented Understudy for
Gisting Evaluation) (presented by Lin \cite{LinRouge04}) can be considered the most common evaluation metric. It compares a \textit{generated summary} (also called \textit{candidate summary}) to one or more gold summaries (also called \textit{reference summaries}). For the comparison, ROUGE-N splits the summaries into n-grams (consecutive sequences of n words) and counts the co-occurrences of these n-grams in generated and gold summaries. Additionally, it is possible to calculate ROUGE-L, which is based on the longest common subsequence of words in the generated summary and the gold summary. With these co-occurrences it is possible to calculate precision, recall and an F-measure for the summaries. With the comparison to a gold summary ROUGE can indicate how much of the content contained in the gold summary is also contained in the generated summary. The possible values range from 0 (low) to 1 (high).

Although ROUGE can be considered the gold standard in evaluating generated summaries, it has several drawbacks. It primarily focuses on word co-occurrences and overlaps (for example noted by Xu et al. \cite{Xu&23QuestionEval}) and cannot discern, for example, synonyms or semantically equal paraphrases (see Egan et al. \cite{Egan&22ShannonScore}). An alternative to ROUGE is BERTScore (presented by Zhang et al. \cite{bert-score}). On a high level view, it works the same way as ROUGE does as it compares which content present in the gold summary is also contained in the generated summary. Different from ROUGE, this content overlap is not calculated by word co-occurrences but by token similarities using the contextual embeddings of a large language model. Assuming that the tokens of synonyms or semantically close words are similar, paraphrased content can also be detected. Just like ROUGE, its values can range from 0 (low) to 1 (high). Other alternative metrics are BARTScore \cite{Yuan&21BARTScore} (which is a trainable metric measuring how likely a summary is generated given the document to summarise) and ShannonScore \cite{Egan&22ShannonScore} (which on an abstract level compares the information contained in the summary with the information contained in the document to summarise). Sometimes even metrics from the evaluation of machine translation (like BLEU \cite{Papineni&Bleu} or METEOR \cite{banerjee-lavie-2005-meteor}) are used. Xu et al. \cite{Xu&23QuestionEval} propose that it may be more viable to evaluate a summaries content based on how good questions generated from the document to summarise can be answered knowing the summary. Min et al. \cite{Min&23FactScore} present a metric focussing on how good the summary reflects the facts of the document to summarise and does not introduce new or wrong information.

\section{Summarisation Task}\label{sec:task}
Our summarisation task is to create \textit{guiding principles} for judgments of the German Federal Court of Justice (in German: Bundesgerichtshof -- subsequently: BGH). The BGH is one of the five highest instance courts in Germany and responsible for the revision of numerous cases in civil and criminal law. As a result, the judgments of the BGH are final and binding for the lower court whose judgment the BGH revises, though the decisions can be subject to a judgment of the German Federal Constitutional Court (but only concerning constitutional matters). Although the judgments are non-binding for other lower courts, the lower courts in general align their judgments in accordance with the decisions issued by the BGH. Hence, the judgments of the BGH significantly influence the legal system in Germany and are of special importance for legal professionals. Besides this practical impact, the judgments of the BGH are ideal for our summarisation task because the BGH delivers plenty of structured and publicly available judgments (520 judgments in 2022). These judgments often (but not always) include guiding principles. In addition, the guiding principles provided by the BGH are rather similar in their structure and style of writing.

In theory, guiding principles are the main sentences reflecting the very essence of the judgment, which judges extract from the text of the judgment  \cite{koebler&24woerter}. Thus, theoretically they are an extractive summary of the judgment. However, in the absence of suitable main sentences or due to the judgment's complexity, the judges of the court summarise the main aspects of the judgment for the guiding principles. In these cases the guiding principles are not a mere extractive summary but indeed an abstractive summary. In our research, we find that a vast majority of guiding principles of the BGH are actually abstractive guiding principles (from our data set of 5081 judgments described below, only 77 have extractive guiding principles).

The guiding principles themselves are non-binding and have no immediate legal effect. However, they serve as guidelines for other, especially lower courts \cite{koebler&24woerter}. In addition, they allow a quick identification of the judgment's main aspects and in this way facilitate the research of suitable judgments and the application of the judgment. As a result, the practical importance of the guiding principles should not be underestimated as in general they are read more often than the judgment. Despite their importance for the German legal system, there is neither an obligation of the court to provide guiding principles nor are there binding rules on how to write them. Naturally, this impedes the determination of the quality of guiding principles from a legal perspective. The judges have a high degree of freedom on how to write guiding principles which leads to difficulties -- for humans and algorithms -- to evaluate the correctness or the quality of generated guiding principles. In the absence of binding rules, C. H. Beck -- one of the leading publishing houses for law literature in Germany -- delivers guidelines for the creation of unofficial guiding principles \cite{beck18Leitsaetze} when publishing judgments without official guiding principles written by judges of the court. These guidelines serve as a basis for the evaluation classes we present in Section \ref{subsec:classes}to assess the quality of generated guiding principles.

In addition, in the majority of cases the judgments of the BGH deliver decisions on legal aspects that have not previously been covered. However, without clear indications in the judgments the models cannot inherently identify novel aspects as they have no historical data. Thus, the creation of guiding principles is a complex task from a technical perspective, too. 

\subsection{Data and Models}
For our following analyses we use 5081 cases of the civil branch of the German Federal Court of Justice from 2003 to 2022.\footnote{We crawl the data from \url{www.rechtsprechung-im-internet.de}. The complete code to our paper is available at \url{https://github.com/bs-000/bgh-ls-llms-ler.git}. }
We split the data into a training set (3556 judgments), a validation set (763 judgments) and a test set (762 judgments). As the legal questions decided in a judgment are written in the dedicated section \textit{reasons for the decision} (in German: \textit{Entscheidungsgründe}), we only consider this section of the written judgment. We further exclude subsection I of the reasons for the decision as in the judgments of the BGH this subsection only gives a brief recap of the previous instance's decision and holds no information about the current judgment to be summarised. These measures limit the inclusion of undesired information and, at the same time, increase the likelihood that the judgment fits the context window of the large language model we use. \textit{In other words, our targeted summarisation task is to generate guiding principles of these judgments from the shortened textual representations of the reasons for the decision.}

In the subsequent sections, we use the \texttt{LeoLM-Mistral-7B} model (in the following: \texttt{LeoLM})\footnote{Available at \url{https://huggingface.co/LeoLM/leo-mistral-hessianai-7b}.} with a context window size of 32k tokens, which is a publicly available open source large language model based on Mistral. The lengths of the different data sets in tokens as tokenized by its tokenizer is shown in Table \ref{tab:length_normal}. The (shortened) reasons for the decision have around 6 800 tokens on average with some outliers exceeding the context window size of the model.\footnote{As these outliers are less then 1\% of our data we do not address this issue in our work and truncate these examples to the context window size.} Guiding principles are much shorter with an average length around 230 tokens, though there are also outliers with a length of around 3k tokens. We did a manual inspection of these outliers and found that they are very rare and concern extremely extraordinary topics which are completely new to the legal system and thus need more clarifications in the guiding principles. As these outliers are substantially different from the other guiding principles, we exclude them from the training set.

\begin{table}[h!]
  \begin{center}
    \caption{Length of the data sets in tokens.}
    \label{tab:length_normal}
    \begin{tabular}{cc|cccc}
      &data set & min	& mean	& max	&std\\
      \hline
      	  &all& 558	&6794.07	&68988&	4495.97\\
        reasons for&train	&670&	6821.48	&68988&	4587.12\\
        the decision&valid	&574	&6587.20	&36885&	3958.16\\
        &test	&558	&6873.29	&48362&	4571.57\\
        \hline
         
        &all&	19	&234.69&	3260&	188.98\\
        guiding&train &	19	&236.83	&3260	&194.53\\
        principles&valid&	23	&233.18	&2109	&188.23\\
        &test	&27	&226.23&	1219	&161.41\\
    \end{tabular}
  \end{center}
\end{table}

\subsection{Automated Summary Evaluation} \label{subsec:metrics}
For the evaluation of our results in the following sections we use the actual guiding principles provided in the judgments as gold summaries to calculate the F-measure of ROUGE-1, ROUGE-2, ROUGE-L and BERTScore. 
For the actual calculation of ROUGE we use our own implementation (also provided in our code) because no existing implementation allows for a suitable sentence preprocessing for German legal texts. 
When calculating BERTScore, we employ the \texttt{microsoft/mdeberta-v3-base} model.

\section{Evaluation classes for Legal Summaries}\label{sec:classes}
For an in-depth evaluation of a legal summary's quality that goes beyond ROUGE or similar metrics, we propose the assessment through different content-related classes. In the following Section \ref{subsec:classes}, we present and compare these classes to existing methods in research. We then show a proof of concept application of our classes to summaries both created by a generative model and a baseline extractive algorithm in Section \ref{sec:poc}.

\subsection{Proposed Evaluation classes}\label{subsec:classes}
For our evaluation classes (see Table \ref{tab:eval_class}) we build on \cite{beck18Leitsaetze} and identify four core aspects of high-quality guiding principles: Language, Pertinence, Completeness, and Correctness. We cover each of these core aspects with at least one class. However, these core aspects can only be evaluated if the text to evaluate is indeed intelligible (class 1). Intelligibility is sufficient to fulfil the requirements of class 1, regardless whether the text is in German or another language. While satisfying the requirements of this class might be unnecessary for purely extractive summarisation methods, generative models might not always produce intelligible results. 

\begin{table}
    \centering
    \caption{Proposed Evaluation classes}
    \label{tab:eval_class}
    \begin{tabular}{c|m{2cm}|m{4.5cm}}
        class & Aspect & Description\\
        \hline
        1 & Intelligibility & Intelligible result\\
        
        2 & Language & Correct use of German language\\
        
        3 & Pertinence & Only necessary information\\
        
        4 & Completeness & Inclusion of every aspect\\
        
        5 & Main Focus & Inclusion of 3/4 of aspects\\
        
        6 & Correctness & No error in legal reasoning\\
        \hline
        7 & Superiority & Superior compared to the original\\
    \end{tabular}
   
\end{table}

Class 2 represents the first of the four key aspects: Language. The fulfilment of this class requires that the text is in German and without grammatical mistakes. In addition, the erroneous use of established legal terms fails this class, as these terms may be used synonymous outside of the professional legal language. 

Class 3 evaluates the pertinence of the summary and thus whether the text contains only content necessary for guiding principles. The determination of the information's necessity is reached through a comparison with the original guiding principles. The additional information must pass a certain threshold and must constitute a significant part of the output; the mere addition of a few unnecessary words is negligible. Naturally, the assessment of necessity can be rather subjective and lead to different results in borderline cases. However, since there are no binding rules on how to write guiding principles we cannot solely evaluate whether the generated guiding principles contain any additional information as this information may actually benefit the purpose of guiding principles. We minimise the discrepancies due to the subjectivity through the assessments of each guiding principle by three independent reviewers which allows us to rely on the majority opinion of our legal experts for the assessment of the quality of the guiding principles (See Section \ref{sec:poc}).

Class 4 reflects the completeness of the summary's content. The text must contain every legal aspect of the original guiding principles. However, since the vast majority of original guiding principles contain several legal aspects the summary may not cover every one of them but only a subset. Even a summary that covers, for example, nine of ten aspects would still fail this class. Therefore, in order to further distinguish the quality of the guiding principles, we add class 5, which indicates whether the summary reflects the main focus of the judgment by covering at least three quarters of all legal aspects contained in the original guiding principles.

Class 6 addresses the core aspect of correctness: it evaluates whether the legal reasoning expressed in the summary is flawless. The text may, for example, flawlessly express a legal obligation but assign it to the wrong party and thus fail the requirements of this class. Since our evaluation covers only a comparative analysis, we exclude errors that occur in parts of the generated summary that are not present in the original guiding principles.

Finally, as there are no binding rules for the creation of guiding principles, we add class 7 to enable reviewers to evaluate whether -- in their opinion -- the generated guiding principles are superior to the original one. The assessment of superiority entails a written mandatory reasoning of the reviewer as it is necessary to comprehend and potentially verify this decision. A reason for superiority could be, for example, the inclusion of information that benefits the aforementioned purposes of the guiding principles.

In addition, we allow the reviewers to write comments for each summary in order to reveal peculiarities that are not covered by our proposed evaluation classes. We intend to use these comments to possibly adjust the evaluation classes for future work.

Our proposed classes allow a conclusive determination of the guiding principles' quality. If the summaries fulfil the requirements of every class, they are superior to the original guiding principles. Once they satisfy the requirements of class 1 to class 6, they can be deemed of high quality, as they should be equal to the original ones. Since these six classes address the core aspects of guiding principles, the failure to comply with just one class can already entail a considerable loss in quality compared to the original guiding principles, rendering them ultimately unsuitable for practical use without an amendment by a legal expert.

Though we design these evaluation classes for guiding principles of German judgments, they are also applicable to a large number of other legal texts and are not limited to the German legal systems. The accurate use of professional legal language, the correctness of the generated text, the exclusion of irrelevant information and the limitation to the relevant content is likely desirable for every legal summary. 

The importance of some of these classes has already been the subject of research. For example, Min et al. \cite{Min&23FactScore} consider the correctness of generated texts (class 6) by fact checking. ROUGE, as the most commonly used evaluation metric, solely focuses on completeness in its original form (as it is \textit{Recall-Oriented}) but may also measure the pertinence if precision of F-score are used. The remaining aspects (language and correctness) are not considered by ROUGE and most other automated metrics (like BERTScore \cite{bert-score} or ShannonScore \cite{Egan&22ShannonScore}) or manual evaluation methods (for examples see Elaraby et al. \cite{Elaraby&24Adding} or Xu et al. \cite{Xu&23QuestionEval}). We argue that all of the presented classes and their corresponding aspects need to be considered to accurately evaluate the quality of generated summaries in the legal field.

\subsection{Proof of Concept Application}\label{sec:poc}
In this section, we show that our evaluation classes work in practice when applied by different legal experts and investigate which classes may lead to discordance between the possible reviewers. Therefore, we now create example summaries for our defined task using two different approaches.  

In the first approach, we create example summaries by simply fine-tuning the \textit{LeoLM} model with  QLora\cite{dettmers&23qlora} on our training data set.\footnote{We train for 10 epochs with a learning rate  of $2^{-4}$, gradient checkpointing and a batch size of 1 due to memory constraints.} Then we generate the guiding principles for our validation set using greedy decoding.  
In order to fully grasp the successful application of the evaluation classes by legal experts, we intend to generate summaries fulfilling, not fulfilling and almost (not) fulfilling the requirements of each class. We assume the model to generate such examples for the classes 3 to 6 without further changes, but most likely not for class 1 and class 2. Therefore, we choose to create some faulty summaries by introducing a coding error which leads to the model sometimes repeating itself until the maximum number of tokens to generate (1000 tokens) is reached. Repetitions in the generated texts are likely to occur sometimes when using generative models.
For the second approach, we choose \textit{LexRank} \cite{Erkan&LexRank04} an extractive algorithm we use to create example summaries with two sentences each for judgments from the validation data set. As this approach merely selects sentences from the original document, we expect the classes 1, 2 and 6 to always be fulfilled by this approach (as the original document is expected to fulfil these classes).

We evaluate the resulting guiding principles with ROUGE and BERTScore. The calculated values can be found in Table \ref{tab:rouge_run_one}. The scores indicate that both approaches create diverse summaries which include positive and negative examples for our evaluation.

\begin{table}[h!]
  \begin{center}
    \caption{Results of the evaluation of the generated guiding principles with ROUGE and BERTScore.}
    \label{tab:rouge_run_one}
    \begin{tabular}{cc|cccc}
      &approach & min	& mean	& max	&std\\
      \hline
       ROUGE-1&LexRank&0.0286	&0.2738	&0.8435	&0.1397\\
       &LeoLM&00.0 &0.1704&0.5581&0.1001\\
       \hline
       ROUGE-2&LexRank&0.0	&0.1084	&0.7042&	0.1309\\
       &LeoLM&0.0	&0.0717&	0.5078	&0.0715\\
       \hline
       ROUGE-L&LexRank&0.0284	&0.2232	&0.7640	&0.1253\\
       &LeoLM&0.0	&0.1608&	0.5520	&0.0973\\
       \hline
       BERTScore&LexRank&0.4783&	0.6524&	0.9149	&0.0612\\
       &LeoLM&0.0	&0.5861	&0.8456&	0.1314\\
    \end{tabular}
  \end{center}
\end{table}

We now apply our defined evaluation classes to a subset of these example summaries. We select 100 random judgments from the data set and compare the example summaries created by both approaches to the original guiding principles of the judgments using the evaluation classes. This manual evaluation is done by five different legal professionals (second to sixth author of this article).\footnote{The detailed instructions for reviewers can be found in our repository.} Four of the reviewers have passed the first legal state examination, while one reviewer has passed its state examination part. The successful passing of the first legal state examination demonstrates that the student has the theoretical and practical skills necessary for the German legal clerkship which is a practical phase that concludes with the second legal state examination. Two of our reviewers also passed the second legal state examination making them fully qualified German lawyers. We make sure that each judgment is evaluated by three of our legal experts to be able measure the degree of agreement between them. The reviewers know that the example summaries are created by a large language model and a baseline algorithm (\textit{LexRank}) but unaware which approach generated which summaries. We measure the overall agreement of our reviewers with Fleiss' Kappa. The agreement scores can be found in Table \ref{tab:aggreement_run_one}. We see that all reviewers have a substantial agreement (as defined by Landis et al. \cite{Landis&77Agreement}) with each other when inspecting all classes for all of the 100 judgments. This indicates that the classes can be applied fairly uniformly by different legal experts.

\begin{table}[h!]
  \begin{center}
    \caption{Agreement scores of the five reviewers calculated with Fleiss' Kappa for all classes.}
    \label{tab:aggreement_run_one}
    \begin{tabular}{c|ccccc}
      Reviewer& A & B & C	& D & E\\
      \hline
      A & 1 & 0.6568 & 0.7141 & 0.6453 & 0.7265\\
      B & 0.6568 & 1& 0.6795 & 0.6567 & 0.7684\\
      C & 0.7141 & 0.6795 & 1 & 0.7075 & 0.7521\\
      D & 0.6453 & 0.6567 & 0.7075 & 1 & 0.6849\\
      E & 0.7265 & 0.7684 & 0.7521 & 0.6849&1\\

    \end{tabular}
  \end{center}
\end{table}

Next, we investigate whether some individual classes show lower agreement scores among the reviewers than others. Therefore, we calculate the agreement scores of the reviewers for the elements of each class separately. The average scores can be found in Table \ref{tab:kappa_per_class}. We observe perfect agreement in class 1, substantial agreement for class 7, moderate agreement for class 5, fair agreement for classes 4 and 6 and slight agreement for classes 2 and 3. 

In general, these scores are lower than the scores observed over all classes. This is reasonable as less data points are used for these calculations (only a seventh part each) thus there is higher uncertainty. The rather low agreement in class 2 is particularly surprising, especially as this even happened for 8 summaries created by \textit{LexRank}, which only extracts sentences from the original judgment. This could be an indicator of a (possibly) low quality of the language in the judgments or there might be some general disagreement on the usage of certain established legal terms among legal experts. When discussing this finding with our reviewers, we find that certain issues are treated differently. An example can be observed with the repetitions of a few words or sentences that we intentionally generate with the language model. Since the repeated phrases are in itself grammatically correct, some reviewers assess them to contain no grammatical errors whilst other reviewers already deem the repetition itself to be a violation of class 2. 

A similar unexpected situation can be observed in class 6 which considers errors in legal reasoning. Of the summaries that reviewers consider to contain errors, a considerable part is again created by \textit{LexRank}. We assume the legal reasoning in these summaries to be correct in the individual sentences and the error to result from considering the concatenated sentences as a whole. Nonetheless, for summaries created by both approaches, we find that our legal experts have slightly differing opinions on what an error in legal reasoning is. According to our reviewers, this is caused mostly by inaccurate wording in the summaries. As an example, the reasoning may lack precision and hence allow an individual interpretation by the reviewer whether the reasoning is correct, which can result in differing opinions. Another point could be the complexity of the inspected judgments: Many of these judgments concern a variety of highly specialised topics where the expertise of our reviewers can differ considerably. Thus, it is conceivable that a reviewer may fail to detect a logical error where the detection requires additional proficiency in the respective area of law.

Class 3 is particularly challenging. This is not surprising as the reviewers are supposed to determine the necessity of additional information for a judgment's guiding principles, if this information is not present in the gold summary. As aforementioned, the assessment of the necessity is rather subjective and likely greatly differs in borderline cases. Thus, a lower degree of agreement is expected.
Class 4 shows fair (and almost moderate) agreement and in class 5 we observe moderate agreement. To gauge the completeness of the content seems to be an easier task for the legal experts. We assume these values to rise if more data points are considered in the analysis and believe these differing decisions to also be (partially) ascribed to the varying nuances in language. 
The agreement in class 7 is close to perfect with the single exception of one reviewer deeming the summary of one judgment to fulfil the requirements of this class. This, just like the agreement in class 1, should be seen with caution: As there is only a small number of examples for one of the two possible answers, these results might not reflect the actual agreement because the decisions are obvious. 

\begin{table}[h!]
  \begin{center}
    \caption{Average Fleiss' Kappa among the five reviewers for the evaluation classes with the number of judgments which (not) fulfilled the classes according to their decisions.}
    \label{tab:kappa_per_class}
    \begin{tabular}{c|cccc}
      class & Fleiss' Kappa & fulfilled & not fulfilled\\
      \hline
      1 & 1.0000 & 591 & 9\\
      2 & 0.1585 & 529 & 71\\
      3 & 0.1107 & 56 & 544\\
      4 & 0.3914 & 118 & 482 \\
      5 & 0.4308 & 273 & 327 \\
      6 & 0.2778 & 442 & 158\\
      7 & 0.7986 & 1 & 599\\

    \end{tabular}
  \end{center}
\end{table}

As an overall result we see that some of our defined classes may be considered more difficult to evaluate by legal experts than others. Especially classes 2 and 3 need further clarifications for the intended reviewers of a summarization tasks. Nonetheless, the overall agreement scores show a substantial degree of agreement between all our reviewers. As of consequence of this we believe that our defined classes can be a viable approach for a qualitative evaluation of legal summaries.

\section{Generating Guiding Principles} \label{sec:round_2}
After defining the classes for an in-depth evaluation of generated guiding principles, we generate high quality guiding principles using the \textit{LeoLM} model -- once with the original judgments and once without an enriched version of the texts. Compared to the simple example summaries we create in the proof of concept application (Section \ref{sec:poc}), we now intend to create high-level summaries of the judgment. We evaluate the resulting summaries using ROUGE and BERTScore, as well as our defined evaluation classes. 

\subsection{Applied Methods}
To create high quality summaries for our task, we fine-tune the \textit{LeoLM} model on our training data set.\footnote{In comparison to the proof of concept application, we do not use QLora and again train for 10 epochs with a learning rate  of $2^{-4}$, gradient checkpointing and a batch size of 1 due to memory constraints.} Then we generate the guiding principles with the model for our test data set.\footnote{We choose to use a different data set than in the proof of concept application to avoid any possible bias.} We use greedy decoding for the generation and generate up to 750 tokens for the guiding principles. We also fine-tune the \textit{LeoLM} model (with the same parameters) on judgments enriched with information about legal entities (based on the work of Leitner et al. \cite{Leitner&20Entities}). They view Legal Entity Recognition (LER) as a Named Entity Recognition (NER) task specialised to the legal domain. Other than the typical entities (for example, people or locations), they also specify legal norms, court decisions, legal literature and case-by-case regulations as entities. We enrich the judgments with this information by inserting tags with the entity's name before and after the entity\footnote{As an example, '§ 125 BGB' becomes '<GS> § 125 BGB </GS>' with 'GS' being the abbreviation for the German word 'Gesetz' which Leitner et al. translate to 'law'.} and adding these tags as special tokens to our language model's vocabulary.\footnote{For the detection of the legal entities, we use the \texttt{bert-german-ler} model available at \url{https://huggingface.co/elenanereiss/bert-german-ler}.} 
As a baseline we also create summaries with a length of two sentences using \textit{LexRank}. 

Compared to existing research on the automated summarisation of German (legal) texts, we create summaries by fine-tuning a state of the art decoder-based pre-trained large language model on our given task. Thus, we have a model trained on a a single (summarisation) task using native German legal judgments. In addition, we enrich the judgments' text with information about legal entities. Although Nguyen et al. \cite{Nguyen&23KeyWords} insert specific keywords in a similar fashion for the summarisation of US legal cases, they use these enriched texts to preselect sentences in a classification task and then generate the final (abstractive) summaries from the preselected sentences, which differs from our work. 

\subsection{Evaluation}
We evaluate the generated summaries using BERTScore and ROUGE. The resulting scores can be found in Table \ref{tab:rouge_run_two}. We see that both fine-tuned models clearly outperform the baseline, with the model fine-tuned on the enriched text (\textit{LerLeoLM}) performing slightly better for each of the metrics. This is a first indicator that the inserted information about the legal entities is a helpful addition for creating summaries.

\begin{table}[h!]
  \begin{center}
    \caption{Results of the evaluation of the generated guiding principles with ROUGE and BERTScore.}
    \label{tab:rouge_run_two}
    \begin{tabular}{cc|cccc}
      &approach & min	& mean	& max	&std\\
      \hline
       &LexRank&0.0123	&0.2597	&0.8333	&0.1316\\
       ROUGE-1&LeoLM&  0.0 & 0.2997& 0.8235& 0.1494\\
       & LerLeoLM& 0.0351& 0.3052& 0.9778&0.1546\\
       \hline
       &LexRank&0.0	&0.0932	&0.8276&	0.1192\\
       ROUGE-2&LeoLM&0.0	&0.1160&	0.7273	&0.1362\\
       & LerLeoLM&    0.0&    0.1231& 0.9355& 0.1483\\
       \hline
       &LexRank&0.0	&0.2098	&0.8333	&0.1161\\
       ROUGE-L&LeoLM&0.0	&0.2463&	0.7353	&0.1331\\
       &LerLeoLM&    0.0351& 0.2488& 0.9333& 0.1417\\
       \hline
       &LexRank&0.4890&	0.6458&	0.9242	&0.0573\\
       BERTScore&LeoLM&0.0	&0.6724	&0.9333&	0.0670\\
       &LerLeoLM&  0.5350& 0.6746& 0.9829& 0.0661\\
    \end{tabular}
  \end{center}
\end{table}

To get further insights into the quality of the generated summaries, we assess a subset with our defined evaluation classes. We select 60 random judgments from the test set and compare the generated summaries of the three different approaches to the original guiding principles of the judgments. This manual evaluation is done by the same five legal professionals as before. They again know that the generated guiding principles are created by a large language model and a baseline model but do not know which generated result was created by which approach. The measured agreement scores can be found in Table \ref{tab:aggreement_run_two}. We can see that there is again a substantial agreement between each pair of our reviewers.

\begin{table}[h!]
  \begin{center}
    \caption{Agreement scores of the five reviewers calculated with Fleiss' Kappa.}
    \label{tab:aggreement_run_two}
    \begin{tabular}{c|ccccc}
      Reviewer& A & B & C	& D & E\\
      \hline
      A & 1 & 0.7052 & 0.7777& 0.6253 & 0.7083\\
      B &  0.7052 & 1& 0.6417 & 0.6886 & 0.7077\\
      C & 0.7777 & 0.6417 & 1 & 0.6548 & 0.7300\\
      D & 0.6253& 0.6886& 0.6548& 1 & 0.7017\\
      E & 0.7083& 0.7077& 0.7300 & 0.7017 &1\\
    \end{tabular}
  \end{center}
\end{table}

We calculate the percentage of judgments which fulfil the requirements of our defined classes to gauge the overall quality of the generated summaries. We consider the requirements of a class to be fulfilled only if at least two of the three reviewers of a judgment deem the requirements of the class fulfilled. By doing so, we intend to reduce the impact of an individual reviewer's subjective assessment and consider a consolidated assessment of all reviewers. The results can be found in Table \ref{tab:classes_run_two}. We see that while \textit{LexRank} performs better in classes 2 and 6, the generative models show higher values in classes 3, 4 and 5. While these two language models seem to perform (almost) equally in classes 3, 4 and 6, the \textit{LerLeoLM} model performs notably better in class 5. This is an indicator that inserting information on legal entities in the judgments might help the model find the relevant content. We also calculate the average number of classes the judgments fulfil according to our reviewers (again assuming that a judgment satisfies the requirements of a class if at least two of the three reviewers of a judgment deem the requirements to be fulfilled). The summaries created by \textit{LexRank} fulfil the requirements of $\approx$ 2.55 classes, the ones generated by the fine-tuned language models $\approx$ 3.38 classes for \textit{LeoLM} and $\approx$ 3.45 classes for \textit{LerLeoLM} respectively. We see again that using the enriched text leads to an improvement in the generated results.

\begin{table}[h!]
  \begin{center}
    \caption{Percentage of judgments fulfilling the given classes averaged over the reviewers for each summary.}
    \label{tab:classes_run_two}
    \begin{tabular}{c|ccccccc}
      class& 1 & 2 & 3	& 4 & 5 & 6 & 7\\
      \hline
      LexRank & 1.00 & 0.97 & 0.02& 0.02& 0.17 & 0.85 & 0.00\\
      LeoLM & 1.00 & 0.92 & 0.23& 0.10& 0.32 & 0.82 & 0.00\\
      LerLeoLM & 1.00 & 0.93 & 0.20& 0.10& 0.42 & 0.80 & 0.00\\
    \end{tabular}
  \end{center}
\end{table}

Our reviewers comment on several aspects of the created summaries. They find that created guiding principles often fail to deliver a comprehensive summary of complex aspects and that some models appear to resort to an extraction of the judgment's first sentence. In our attempt to test this finding with an automated comparison, we could not verify it for exact matches. Still, it is possible that the generated guiding principles only slightly vary from the first sentences of the judgments. In addition, the different approaches sometimes appear to address similar aspects of the judgments that are not covered by the original guiding principles and, in this case, utterly fail the summarisation task. Furthermore, the generated guiding principles are not free from the issue of hallucination. For example, the generated guiding principles may cite certain decisions of the BGH or law literature that simply do not exist or are unrelated to the judgment. 

Despite these apparent flaws, several generated guiding principles seem to have a high quality that is on par with the original guiding principles and in very few cases surpass them (class 7) according to individual reviewers. One reviewer deems generated guiding principles as superior because they are almost identical to the original ones but also cite the relevant provision, which the reviewer finds beneficial for the purpose of guiding principles as aforementioned in Section \ref{sec:task}. However, we notice that even good examples of the generated guiding principles often contain certain flaws that ultimately render them unsuitable for a use in practice without further revision. For example, even generated guiding principles covering every aspect of the original guiding principles may suffer from a poor choice of words that may not necessarily be wrong but appear unnatural to the reviewer. In parallel, the paraphrasing leads to the use of certain terms -- which may be synonymous in a non-legal context -- instead of an established, non-interchangeable legal term.

When inspecting several judgments with summaries that were deemed good or bad examples by our reviewers, we notice that the comparative quality of the generated guiding principles depends on both the length and the amount of legal areas covered by the original guiding principles. In this regard, short guiding principles that cover only a very specific area of law tend to yield better results. Interestingly, we notice that the results appear to be unaffected by the complexity of that area of law and, for example, short guiding principles for a judgment in the rather complex area of insolvency law yield good results. However, the models appear to struggle with the creation of guiding principles that consist of a combination of procedural law and substantive law, which oftentimes leads to considerable loss in quality. We suspect that this might be due to the models' tendency to only include aspects regarding substantive law. 

In summary, the manual evaluation revealed rather promising results, as the quality of several generated guiding principles is high or even on par with the original ones. However, the evaluation also shows that the majority of the generated guiding principles contains severe flaws and thus even the best model is not fit for an application in practice without considerable improvements.

\section{Discussion and Limitations} \label{sec:discussion}
We now discuss our results and the limitations of our work. First, we consider the proposed evaluation classes and then discuss our proposed summarisation method.

Regarding the evaluation classes and the corresponding proof of concept application (Section \ref{sec:classes}) we observe some classes whose requirements are (not) fulfilled only by a very small number of examples. Especially classes 1 and 7 show little examples of their requirements not being fulfilled (class 1) and being fulfilled (class 7). We inspect the summaries not satisfying the requirements of class 1 and find that all of these summaries were empty texts, meaning that the language model generated no text at all. Since there is no text, it is not surprising that they cannot fulfil class 1. Additionally, in our analysis of the summaries in Section \ref{sec:round_2}, there are no summaries at all which do not satisfy class 1. We do not observe possible other errors like, for example, meaningless character sequences. Therefore, further experiments are needed, maybe with synthetically created faulty summaries or a generative model with significant errors in it's training to gauge whether class 1 can be uniformly evaluated by different legal experts when considering such erroneous texts. Nonetheless, we believe class 1 to be one of the easier classes to evaluate and do not expect a large disagreement between reviewers. Class 7 is only rated to be fulfilled once in our proof of concept application in Section \ref{sec:classes} and only twice in our analysis in Section \ref{sec:round_2} by individual reviewers -- but never by the majority of all reviewers of one judgment. Creating summaries of this class is extremely difficult and we expect the evaluation of this class to be equally challenging. Therefore, the results for this class must always be inspected critically. We try to reduce the uncertainty of this class' assessment by averaging over the decisions of the three reviewers of each summary when evaluating our high quality summaries in Section \ref{sec:round_2}. 
We also consider whether the assessment of the classes is too obvious for our reviewers which might lead to an artificially high agreement. As we can only report substantial agreements among each pair of our reviewers and for individual classes, we do not believe this to be the case. If the decisions were too obvious, the agreement scores should be higher. Only class 1 shows a perfect agreement between all reviewers, which may be due to the reasons described above.

We also consider the possibility of our reviewers to be biased regarding summaries created by generative models compared to our baseline extractive algorithm. To prevent this, we ensure that the reviewers never know which summary is created by which approach. The possibility still remains that reviewers may be able to differentiate the resulting summaries by some features of the texts themselves. Reviewers might be able to differentiate the generated summaries by the use of professional legal language in the texts. We expect the summaries extracted from the original judgment to always be of fitting professional language and contain no logical errors (class 2 and class 6), while the texts generated by the language models might sometimes be lacking. This assumption proves itself to be wrong when we inspect the annotated data. Our baseline model performs only slightly better then the generative models in these classes and by far not perfect. Another point used for the differentiation might be the length of the created summaries. The baseline method always produced summaries with a length of two sentences, while our generative models could create a much larger number of sentences. In our analysis in Section \ref{sec:classes}, the generated summaries were on average indeed much longer and could therefore be differentiated using the length. Since we do not make any differentiation between the two approaches when evaluating the proof of concept, we do not consider this to be a problem. However, when we analyse the high quality summaries in Section \ref{sec:round_2}, we inspect the results individually for each approach. When we inspect the average lengths of the summaries created by our three approaches, we find that all of them create summaries with a length of two sentences on average and that around one third of all summaries created by the generative models consists of two sentences. Additionally, since the reviewers do not know that the baseline approach always creates summaries consisting of exactly two sentences, using the length of the generated summaries for a differentiation between the three approaches appears extremely difficult. 

Another point to discuss are the improvements of legal summarisation methods made in our work compared to the baseline. When inspecting the results of the automated metrics (ROUGE and BERTScore, see Table \ref{tab:rouge_run_two}), the difference between the baseline and the generative models for the average scores is $\approx$ 0.04 for for all metrics. This does not appear to be much on a first glance. Our manual evaluation of the results (summarised in Table \ref{tab:classes_run_two}) however show a notable improvement between the baseline and the generative models in the three classes that the metrics are likely to measure (classes 4 and 5 measure how much of the information in the gold summary is also contained in the created summary and class 3 measures how much content in the created summary is not contained in the gold summary). From an abstract point of view, these classes intend to measure the same as ROUGE (or BERTScore) Recall and Precision. We indeed find low positive correlations between ROUGE / BERTScore Recall and class 4, medium positive correlations between ROUGE / BERTScore Recall and class 5, as well as medium positive correlations  between ROUGE/BERTScore Precision and class 3 when calculating Spearman correlations (following the interpretation of Cohen \cite{Cohen88Cor}).  In fact, if one were to assume class 4 or class 5 to be the (ideal) Recall result and class 3 the (ideal) Precision result of ROUGE or BERTScore, we get a much larger difference between our approaches and the baseline. We believe the smaller measured improvements calculated by the automated metrics to be caused by false positives and false negatives in the matching of words (ROUGE) and content (BERTScore) -- which is a common problem of ROUGE we describe in Section \ref{subsec:metrics} and maybe also a problem of BERTScore. When we check our work against other results in literature, only Glaser et al. \cite{Glaser&21German} summarise a comparable data set and achieve scores similar to ours. 

Lastly, we address the problem of hallucinated text created by generative models. Our reviewers report that many of these summaries include references to legal literature or judgments which do not exist. For one of the created summaries, this is the only thing which prevents it from being as good as the original guiding principles. The hallucinations are a serious problem for creating useful and meaningful legal summaries. We also analyse whether using the text enriched with legal entities -- which so to speak highlights the correct references in the document to summarise -- reduces the hallucinations for these entities. This does not seem to be the case. Although the model trained with the enriched texts generates summaries containing more legal entities, it apparently does not create more correct entities on average. We assume a generated entity to be correct if it is also contained in the document to summarise. Therefore, it might be possible that we missed some correct entities (for example, a correct citation of a related case) if it is not also cited in the original judgment. We believe this to be the case only for a very small number of entities, if any at all.

\section{Conclusion} \label{sec:conclusion}
In this paper we consider the problem of creating guiding principles for judgments of the German Federal Court of Justice using a decoder-based large language model. For a proper evaluation of the quality of the generated results -- other than with existing automated metrics (like ROUGE) -- we present a set of evaluation classes for a manual evaluation to assess the language, pertinence, completeness and correctness of created summaries. 
We then fine-tune the decoder-based generative model on our task using the original judgments' texts and the judgments' texts enriched with information about legal entities. Our evaluation of the results show notable improvements compared to the baseline. We also find that the enhancement of the texts strengthens the model's ability to identify the document's content relevant for the summarisation task. 

For future improvements of our evaluation classes, we intend to address the classes with a considerable disagreement between the reviewers, such as class 2 and class 3. 
In addition, class 7 appears to be impractical for the review process as the reviewer's assessment is highly subjective and requires a certain expertise to properly evaluate the superiority. Thus, we will resort to comments for a manual comparative analysis of the generated guiding principles. We also intend to explore different options to aid the evaluation process of the reviewers with an automated evaluation or (partially) automating the process. 

Regarding the summarisation task, we plan to further explore the potential of using legal entities. 
Another aspect we intend to address in future work is the length of the texts. 
Due to the small number of judgments exceeding our model's context window size in our data set (<1\%), we neglect this aspect and simply truncate the judgments, which potentially leads to a loss of important information. 
Finally, our data set seems to at least partially overlap with the data set used by Glaser et al. \cite{Glaser&21German} and we intend analyse the applicability of our approach to this data set.


\begin{acks}
Computational resources were provided by the German AI Service Center WestAI.
\end{acks}

\bibliographystyle{ACM-Reference-Format}
\bibliography{main}


\end{document}